\documentclass{article}

\PassOptionsToPackage{numbers, compress}{natbib}

     \usepackage[preprint]{neurips_2023}

\usepackage[utf8]{inputenc} 
\usepackage[T1]{fontenc}    
\usepackage{hyperref}       
\usepackage{url}            
\usepackage{booktabs}       
\usepackage{amsfonts}       
\usepackage{nicefrac}       
\usepackage{microtype}      
\usepackage{xcolor}         
\usepackage{graphicx}

\usepackage{wrapfig}
\usepackage{lscape}
\usepackage{rotating}
\usepackage{color}
\definecolor{orange}{RGB}{255,127,0}
\definecolor{brown}{RGB}{150,70,0}
\definecolor{green}{RGB}{127,255,127}
\definecolor{darkgreen}{RGB}{0,127,0}
\definecolor{blue}{RGB}{127,127,255}
\definecolor{lightblue}{RGB}{150,150,255}
\definecolor{darkblue}{RGB}{0,0,127}
\definecolor{red}{RGB}{255,90,90}
\definecolor{violet}{RGB}{200,110,170}
\definecolor{grey}{RGB}{127,127,127}
\definecolor{pink}{RGB}{255,180,180}

\ifdefined\sidenotes

\newcommand{\sn}[1]{\textcolor{blue}{${\leftarrow}\hspace{-3pt}{\bullet}$\marginpar{\textcolor{blue}{${\leftarrow}\hspace{-3pt}{\bullet}$}\tiny{\textcolor{blue}{#1}}}}}

\newcommand{\snJHO}[1]{\textcolor{violet}{${}\hspace{-3pt}{\bullet}$}\marginpar{\textcolor{violet}{${\leftarrow}\hspace{-3pt}{\bullet}$}\tiny{\textcolor{violet}{Jose: #1}}}}

\newcommand{\snRyan}[1]{\textcolor{darkblue}{${}\hspace{-3pt}{\bullet}$}\marginpar{\textcolor{darkblue}{${\leftarrow}\hspace{-3pt}{\bullet}$}\tiny{\textcolor{darkblue}{Ryan: #1}}}}

\newcommand{\snHan}[1]{\textcolor{green}{${}\hspace{-3pt}{\bullet}$}\marginpar{\textcolor{darkblue}{${\leftarrow}\hspace{-3pt}{\bullet}$}\tiny{\textcolor{darkblue}{Ryan: #1}}}}
                
\else

\newcommand{\sn}[1]{}
\newcommand{\snJHO}[1]{}
\newcommand{\snRyan}[1]{}
\newcommand{\snHan}[1]{}
                
\fi

\title{Revealing the structure of language model capabilities}

\author{%
  Ryan Burnell\\
  Leverhulme Centre for the Future of Intelligence,\\
  University of Cambridge\\
  The Alan Turing Institute\\
  \textit{rburnell@turing.ac.uk} \\
  \And
  Han Hao\\
  New Mexico State University\\
  \AND
  Andrew R. A. Conway\\
  New Mexico State University\\
  \And
  Jose Hernandez Orallo\\
  Universitat Politecnica de Valencia\\
  Leverhulme Centre for the Future of Intelligence, \\
  University of Cambridge\\
  Centre for the Study of Existential Risk, \\
  University of Cambridge
}

\begin{document}

\maketitle

\begin{abstract}
  Building a theoretical understanding of the capabilities of large language models (LLMs) is vital for our ability to predict and explain the behavior of these systems. Here, we investigate the structure of LLM capabilities by extracting latent capabilities from patterns of individual differences across a varied population of LLMs. Using a combination of Bayesian and frequentist factor analysis, we analyzed data from 29 different LLMs across 27 cognitive tasks. We found evidence that LLM capabilities are not monolithic. Instead, they are better explained by three well-delineated factors that represent reasoning, comprehension and core language modeling. Moreover, we found that these three factors can explain a high proportion of the variance in model performance. These results reveal a consistent structure in the capabilities of different LLMs and demonstrate the multifaceted nature of these capabilities. We also found that the three abilities show different relationships to model properties such as model size and instruction tuning. These patterns help refine our understanding of scaling laws and indicate that changes to a model that improve one ability might simultaneously impair others. Based on these findings, we  suggest that benchmarks could be  streamlined by focusing on tasks that tap into each broad model ability.

\end{abstract}

\section{Introduction} \label{intro}

\snJHO{We only mention scaling laws once in the main paper, but it's also mentioned in the abstract. I think the introduction lacks motivation, a lot. Just that we don't understand LLMs and the third paragraph directly to the theories of intelligence in psychometrics. And what are the contributions of the paper?}
\snRyan{The contributions are in the abstract, I don't think we need to rehash them here! I've tried to rework this section again.}
The capabilities of Large language models (LLMs) are growing broader every year. Yet our understanding of the these capabilities remains remarkably narrow. Because these systems are so general, it has proved incredibly challenging to create benchmarks that can provide an informative account of their capabilities and limitations. In an attempt to do so, researchers have developed large benchmarks such as Big-BENCH and HELM \cite{srivastava2022,liang2022} that test the performance of LLMs on a variety of different tasks. These benchmarks are an important step in the right direction, but they currently lack both explanatory and predictive value---although they might be able to tell us that a model performs poorly on a particular set of tasks, they cannot tell us why the model struggled nor accurately predict how it will behave for a new, untested task.

Part of the problem is that the underlying abilities that enable LLMs to perform well across so many tasks are poorly understood \cite{taylor2021, burnell2022, burnell2023}. We do not yet have a clear idea of how to meaningfully characterize these abilities, how they are structured, or how they are related to one another. We also lack a full understanding of how the properties of a model affect specific aspects of its abilities. Addressing these gaps is vital for our ability to explain and predict the performance of LLMs across diverse deployment scenarios---after all, if we do not understand the capabilities of these models, it is difficult to anticipate their behavior \cite{srivastava2022}. Moreover, a strong understanding of LLM capabilities would provide more theoretically grounded ways of creating evaluation benchmarks, potentially making these benchmarks both more robust and more efficient. For all these reasons, we aim in this paper to build a deeper understanding of the capabilities of LLMs by drawing on empirical techniques from cognitive science. 

Understanding the nature of intelligence has long been a focus of cognitive science. Over the past century, psychometric theories of human intelligence, such as the Cattell-Horn-Carroll theory \cite{cattell1978, carroll1997, mcgrew2009}, have demonstrated that a great deal of the variance among people's cognitive performance on different tasks can be explained by a relatively small set of latent factors that represent broad cognitive abilities. These separable but interrelated abilities are considered to be essential components of human cognition \cite{sternberg2013, conway2015}. These abilities include the ability to solve complex novel problems, known as \textit{fluid intelligence}, the ability to comprehend and produce written language, known as \textit{reading and writing ability}, and the breadth and depth of one’s acquired knowledge about the world, known as \textit{domain knowledge}.

Research into the structure of human cognition has provided a number of benefits. At a high level, this work has allowed us to build a theoretical understanding of the mechanisms that underpin human behavior. At a more practical level, this understanding as allowed us to build robust tests of cognitive abilities \cite{kaufman2014} that enable the accurate prediction of a wide range of life outcomes, including academic achievement and job performance \cite{conway2015, flanagan2013}. This understanding also makes it possible to identify and explain cognitive deficits or learning disabilities in specific individuals, which in turn helps inform interventions and treatments \cite{flanagan2010, miller2008}.

In a similar way, there are likely to be many benefits to building an understanding of the structure of large language model capabilities. At a high level, such an understanding would help us figure out how these ``black box'' models function and how they process information \cite{taylor2021}. More practically, this understanding would help inform the construction of evaluation benchmarks that can identify the strengths and limitations of these models, enabling more robust predictions about their behavior on a wide range of tasks. Finally, if we can identify deficiencies in the capabilities of specific models, we can better target efforts to address those deficiencies.

One useful way to build this understanding would be to utilize factor analysis techniques, which can be used to extract latent factors that explain variance across different tasks. The data-driven, ``bottom-up'' nature of these techniques makes them an excellent starting point for empirical investigations into the structure of language model capabilities. This approach also fits with research in cognitive science, which has long used factor analysis to build an understanding of the structure of intelligence in humans and animals.

Until recently, the relatively small number of language models capable of complex behavior had rendered this approach infeasible. However, the recent explosion in the number of LLMs of various scales, architectures, and training regimes now makes it possible to collect data from a varied population of models. In addition, over the past few years, large test benchmarks designed to evaluate the broad capabilities of LLMs have been developed, such as BIG-bench and HELM \cite{srivastava2022, liang2022}. These benchmarks provide a standardized way of testing LLMs across a variety of tasks, many of which were designed to test specific cognitive abilities. With data from this new population of LLMs across a range of these benchmark tasks, it is now becoming feasible to use an individual differences approach to investigate the structure of language model capabilities.

What hypotheses, though, might we make about the structure of language model capabilities? It seems likely that the capabilities of LLMs are multifaceted. After all, we can find this multifaceted pattern across a number of different species, including humans, pigeons, and chimpanzees \cite{flaim2020, schneider2012}. This idea also fits with the existing literature on LLMs---for example, there is  evidence that new abilities can suddenly “emerge” at particular scales \cite{wei2022, srivastava2022}, which is consistent with the idea that certain abilities can be dissociated from one another. What is more uncertain is how these abilities are structured. For example, can the ability of language models to comprehend language be largely explained by a single broad ability, as is the case in humans \cite{schneider2012}? Or are there multiple distinct abilities involved? Similarly, a range of studies have investigated the ability of LLMs to perform various kinds of reasoning, including quantitative reasoning, deductive reasoning, and commonsense reasoning \cite{rajani2019, clark2020, lewkowycz2022}. But it remains unclear whether these different kinds of tasks rely on a set of distinct abilities or rather on a single, underlying reasoning ability.

Once we have identified the broad abilities of LLMs, we can begin to examine how model properties affect these different abilities. For example, a great deal of work has already been done to examine how the scale of a model affects its performance. In general, there is evidence increasing model size tends to improve performance, a pattern often referred to as "scaling laws" \cite{srivastava2022,hoffmann2022}. This pattern mirrors findings from cognitive science showing relationships between brain size and intelligence \cite{healy2007,logan2018}. But at the level of specific tasks, the findings from the LLM literature are more complex and difficult to interpret. For example, data from BIG-bench suggests that model scale is closely related to performance on some tasks, but bears no relationship to performance on other tasks \cite{srivastava2022}. What are we to make of this pattern? One interpretation is that these two sets of tasks rely on different abilities, and that these abilities have differing relationships with model size. In order to disentangle these complex patterns, we need to identify the underlying abilities involved in different tasks and examine the relationships between these abilities and various model properties. If we can do that, we may be better able to predict how changes to model properties might affect performance in specific domains. 

\section{Methods}

To investigate the structure of large language model capabilities, we drew on a large dataset released by the creators of the HELM benchmark \cite{liang2022} that includes performance data from a variety of LLMs. We used this dataset to extract latent factors representing the broad abilities of LLMs, and then examined the relationships between those capabilities and model properties, such as model size and instruction tuning. The full dataset and analysis code are available on Github \url{https://github.com/RyanBurnell/revealing-LLM-capabilities/}.

\subsection{HELM data} \label{HELM}
Here, we analysed benchmark data from 29 LLMs tested on the HELM benchmark \cite{liang2022}. The models vary in size, release date, training data, and architecture. These include variants of GPT-3 and InstructGPT as well as models from Cohere, Anthropic, Google, Microsoft/NVIDIA, and AI21. Models were only included in the dataset if they contained missing data on fewer than 3 of the included tasks. A full list of the models and their characteristics can be found in Table~\ref{tab:models}. These characteristics were drawn from a survey of LLMs \cite{zhao2023} as well as the papers introducing each model. Instruction tuning (IT) refers to the process of fine tuning the model to respond to instructions and answer questions. Reinforcement learning from human feedback (RLHF) refers to the process of fine tuning the model to provide responses that are liked by humans.

\begin{table}[!h]
\centering
    \begin{tabular}{lrrrrr}
        \toprule
        Model & Size (B) & Total Tokens & Release Date & IT & RLHF \\
        \midrule
        Anthropic-LM v4-s3 & 52 & 4.00E+11 & 01/12/2021 & 1 & 1 \\
        BLOOM & 176.00 & 3.66E+11 & 01/07/2022 & 0 & 0 \\
        Cohere Command beta & 52.4 & ? & 03/01/2023 & 1 & ? \\
        Cohere Command beta & 6.1 & ? & 03/01/2023 & 1 & ? \\
        Cohere large v20220720 & 13.1 & ? & 20/07/2022 & 0 & 0 \\
        Cohere medium v20220720 & 6.1 & ? & 20/07/2022 & 0 & 0 \\
        Cohere medium v20221108 & 6.1 & ? & 08/11/2022 & 0 & 0 \\
        Cohere small v20220720 & 0.41 & ? & 20/07/2022 & 0 & 0 \\
        Cohere xlarge v20220609 & 52.4 & ? & 09/06/2022 & 0 & 0 \\
        Cohere xlarge v20221108 & 52.4 & ? & 08/11/2022 & 0 & 0 \\
        GPT-3-ada & 2.7 & 3.00E+11 & 01/05/2020 & 0 & 0 \\
        GPT-3-babbage & 6.7 & 3.00E+11 & 01/05/2020 & 0 & 0 \\
        GPT-3-curie & 13 & 3.00E+11 & 01/05/2020 & 0 & 0 \\
        GPT-3-davinci & 175 & 3.00E+11 & 01/05/2020 & 0 & 0 \\
        GPT-J & 6 & 4.00E+11 & 04/06/2021 & 0 & 0 \\
        GPT-NeoX & 20 & 4.00E+11 & 14/04/2022 & 0 & 0 \\
        InstructGPT-text-ada-001 & 2.7 & 3.00E+11 & 27/01/2022 & 1 & 1 \\
        InstructGPT-text-babbage-001 & 6.7 & 3.00E+11 & 27/01/2022 & 1 & 1 \\
        InstructGPT-text-curie-001 & 13 & 3.00E+11 & 27/01/2022 & 1 & 1 \\
        InstructGPT-text-davinci-002 & 175 & 3.00E+11 & 27/01/2022 & 1 & 1 \\
        InstructGPT-text-davinci-003 & 175 & 3.00E+11 & 28/11/2022 & 1 & 1 \\
        J1-Grande v1 & 17 & ? & 01/08/2021 & 0 & 0 \\
        J1-Grande v2 beta & 17 & ? & 01/08/2021 & 0 & 0 \\
        J1-Jumbo v1 & 178 & 3.00E+11 & 01/08/2021 & 0 & 0 \\
        J1-Large v1 & 7 & 3.00E+11 & 01/08/2021 & 0 & 0 \\
        OPT & 175 & 1.80E+11 & 22/12/2022 & 0 & 0 \\
        OPT & 66 & 1.80E+11 & 22/12/2022 & 0 & 0 \\
        TNLG v2 & 530 & 3.39E+11 & 28/01/2022 & 0 & 0 \\
        TNLG v2 & 6.7 & 3.39E+11 & 28/01/2022 & 0 & 0\\
        \bottomrule
    \end{tabular}
    \caption{The 29 models included in the analysis, the model parameter count (billions), the total number of tokens trained on, the release data of the model and whether the model underwent instruction tuning (IT) and/or reinforcement learning from human feedback (RLHF). "?" denotes that the characteristic could not be determined.}
    \label{tab:models}
\end{table}

\subsection{Task selection} \label{taskSelection}
The HELM dataset contains performance data from 34 different tasks. However, several of these tasks contained missing data for a number of models. Several other tasks lacked a clear cognitive basis (e.g., the data imputation task). In addition, the NaturalQuestions task includes two very similar variations (open book and closed book). Because we were limited by the number of models included in the HELM dataset, we removed these tasks to minimize the number of parameters to be estimated by the factor analysis models. The full list of included tasks can be found in Figure ~\ref{fig:loadings}, and more details about each task can be found in the HELM benchmark paper \cite{liang2022}. Note that we included two versions of the synthetic reasoning task---one based on abstract symbols (A), and one based on natural language (NL).

\subsection{Task demand coding} \label{demandCoding}
To assist with the interpretation of any extracted factors, a cognitive science expert (the first author) annotated each task according to the primary cognitive ability tested by the task. The task annotations can be found in Figure ~\ref{fig:loadings}. Note that MMLU was annotated as mixed because it includes a range of sub-tasks, some of which rely primarily on domain knowledge, and others that rely primarily on fluid reasoning \cite{hendrycks2021}.

\section{Results}

\subsection{Task correlations}
The simplest way to investigate how performance varies across different tasks is to examine the correlations between the tasks across the population of LLMs. Doing so, we find a largely positive correlation manifold, with a mean correlation of \textit{r} = 0.56, (\textit{Med} = 0.6, see Appendix \ref{appendixA} for the full correlation matrix). In general, the correlations were strongest between the tasks that were annotated as relying on the same cognitive abilities, which provides some initial evidence for the idea that these tasks fit together conceptually. But given the complexity of the correlation matrix, and because each task might require more than one ability, looking at the raw correlations does not necessarily tell the full story.
                                    
\subsection{Factor analysis} \label{fa}
To build a better model of the abilities that can explain performance across a range of tasks, we draw on factor analytic techniques commonly used to understand the structure of cognition in humans and non-human animals \cite{fabrigar2012, rosseel2012}. These techniques are designed to extract latent variables based on individual differences in performance across a range of different tasks. Studies using frequentist factor analytic typically employ larger samples than it would be possible to obtain from modern LLMs \cite{maccallum1999}, so we also employed Bayesian factor analysis as a robustness check.

Because we lacked a clear a priori theoretical model of the structure of LLMs, we first aimed to determine the number of factors that best explain the data. To do so, we used the Hull method \cite{lorenzo-seva2011}, which aims to balance model fit and model simplicity. This method suggests that a 3-factor solution is most appropriate (see Appendix \ref{appendixB} for more details).

Based on this result, we conducted a frequentist maximum likelihood exploratory factor analysis (EFA) assuming 3 factors and using the oblimin rotation method (an oblique rotation method that allows the factors to be correlated) \cite{rosseel2012}.\footnote{Using an orthogonal rotation method such as varimax produces very similar results.}. As expected given the small sample, the model fit statistics were somewhat poor (CFI = 0.70, TLI = 0.61, RMSEA = 0.26). Nonetheless, the analysis produced three clear factors, each of which explains a substantial proportion of the variance across the different LLMs (see Table \ref{tab:variance}).

\begin{table}[h]
\centering
\begin{tabular}{lccc}
\toprule
    & F1 & F2 & F3 \\
\midrule
Proportion var. explained & 0.33 & 0.31 & 0.17 \\
Cumulative var. explained & 0.33 & 0.64 & 0.82 \\

\bottomrule
\end{tabular}
\caption{Variance explained by each of the three factors from the frequentist EFA}
\label{tab:variance}
\end{table}

\begin{figure}[!b]
\centering
\includegraphics[width = 1\textwidth]{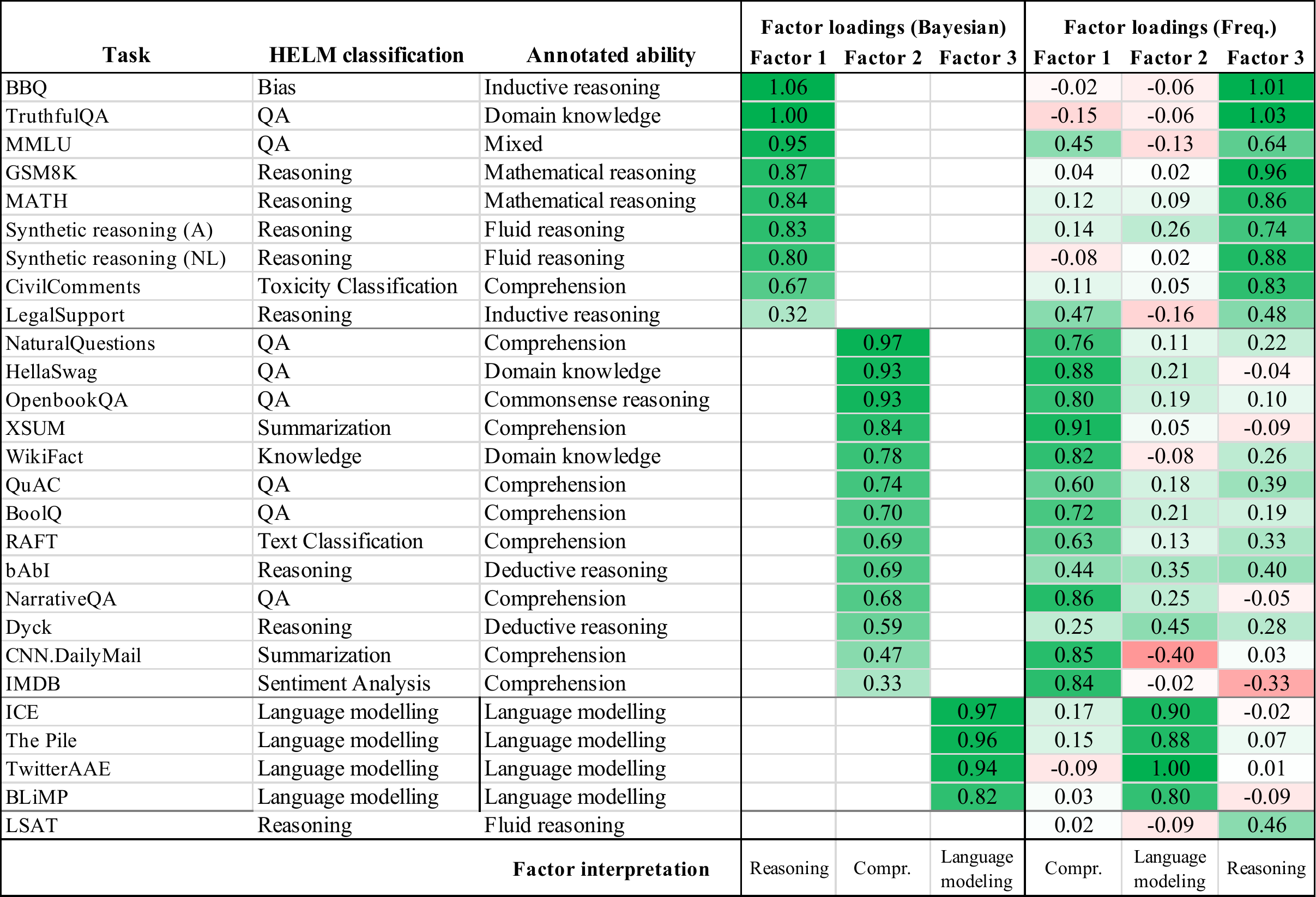}
\caption{Task annotations and factor loadings for each task from the Bayesian factor analysis (left) and the frequentist factor analysis (right). Darker greens represent stronger positive loadings, darker reds represent stronger negative loadings. Note that the Bayesian method proposed by Conti et al. \cite{conti2014} only calculates factor loadings with the assigned factor.}
\label{fig:loadings}
\end{figure}

Based on the task annotations, a relatively clear picture emerged of the three extracted factors. The right-most three columns in Fig. \ref{fig:loadings} display the factor loadings for each task on each of the three factors, which can be broadly interpreted as capturing comprehension, language modelling, and reasoning, respectively.  

We found that tasks testing comprehension generally loaded strongly on factor 1. These include NaturalQuestions (which involves answering questions about a passage from from wikipedia provided in the context window),  XSUM (which involves summarizing articles from the BBC provided in the context window), and NarrativeQA (which involves answering questions about a book or movie script provided in the context window). 

Tasks explicitly testing the language modeling ability of the model through next-token prediction of the model (as measured by bits-per-byte) tended to load strongly on factor 2. These include ICE (which tests next-token prediction on the The International Corpus of English), TwitterAAE (which tests next-token prediction on African-American-aligned and White-aligned English tweets), and The Pile (which tests next-token prediction on an array of English texts from The Pile dataset).

Reasoning tasks generally loaded strongly on factor 3. These include GSM8K (a mathematical reasoning task), both synthetic reasoning tasks (pattern matching tasks designed to measure fluid reasoning), and BBQ (an inductive reasoning task that is used to test for bias in a system). 

Several other notable patterns emerged from the loadings. First, the two tasks designed to measure cognitive biases (BBQ and TruthfulQA) loaded strongly on the reasoning factor. This could be because in order to perform well in these tasks, models need to reason about the correct answer ``in the moment'' and avoid relying on stored knowledge that may lead to an incorrect response. Second, the two deductive reasoning tasks (bAbI and Dyck) loaded only weakly on the reasoning factor, while the commonsense reasoning task (HellaSwag) loaded on the comprehension factor. This may indicate that the reasoning factor primarily represents the model's ability to engage in complex forms of inductive reasoning---a possibility worthy of further investigation.

Finally, we found that the three extracted factors were moderately correlated with one another (see the top 2 rows of Table \ref{tab:correlations}). This pattern fits with findings from cognitive science demonstrating that human cognitive abilities tend to be inter-correlated, despite the fact that they can be dissociated from one another---this is thought to be because complex tasks require tend to require multiple abilities working in combination.\cite{kovacs2016}. 

\begin{table}[h]
\centering
\begin{tabular}{lrrr}
\toprule
 & F1 (Comprehension) & F2 (Language modeling) & F3 (Reasoning)\\
\midrule
F2 (Language modeling) & 0.43 {[}0.09, 0.68{]} &  &  \\
F3 (Reasoning) & 0.51 {[}0.19, 0.73{]} & 0.22 {[}-0.14, 0.53{]}&  \\
Log model size & 0.70 {[}0.46, 0.85{]} & 0.49 {[}0.16, 0.72{]} & 0.51 {[}0.19, 0.73{]}\\
Instruction tuning & 0.23 {[}-0.13, 0.54{]} & -0.50 {[}-0.72, -0.17{]} & 0.44 {[}0.11, 0.69{]}\\
Training length (tokens) &  -0.02 {[}-0.47, 0.44{]} & 0.11 {[}-0.36, 0.54{]} & 0.08 {[}-0.38, 0.52{]}\\
\bottomrule
\end{tabular}
\caption{Pearson correlations (with 95\% confidence intervals) between extracted abilities and model characteristics. Log model size was used due to the exponential distribution of the model sizes in the data.}
\label{tab:correlations}
\end{table}

\subsection{Bayesian factor analysis}
Because of the relatively small population of subjects in our dataset and the somewhat poor fit of the model, some caution is needed when interpreting the results from this analysis. Therefore, to test the robustness of our findings, we next conducted a Bayesian Factor analysis using the method proposed by Conti and colleagues \cite{conti2014} because Bayesian methods are much more robust to small sample sizes \cite{mcneish2016}. This analysis method determines the optimal factor solution as part of the modeling process, then assigns each task to a single factor and calculates the factor loadings with that factor. In general, Bayesian EFA is relatively more conservative and data-driven than frequentist EFA techniques.

The results of the Bayesian factor analysis align closely with the results from the frequentist analysis, which gives us confidence in the reliability of the findings. As Fig. \ref{fig:loadings} shows, a clear 3-factor solution emerged, with factor loadings that largely match those from the frequentist analysis. The most notable exception was the LSAT task, which was not assigned to any of the three factors by the Bayesian analysis but loaded moderately with the reasoning factor in the frequentist analysis (which makes sense, given that it was designed as a fluid reasoning task).

\subsection{Model factor scores}
\begin{figure}[!h]
\centering
\includegraphics[width = 1\textwidth]{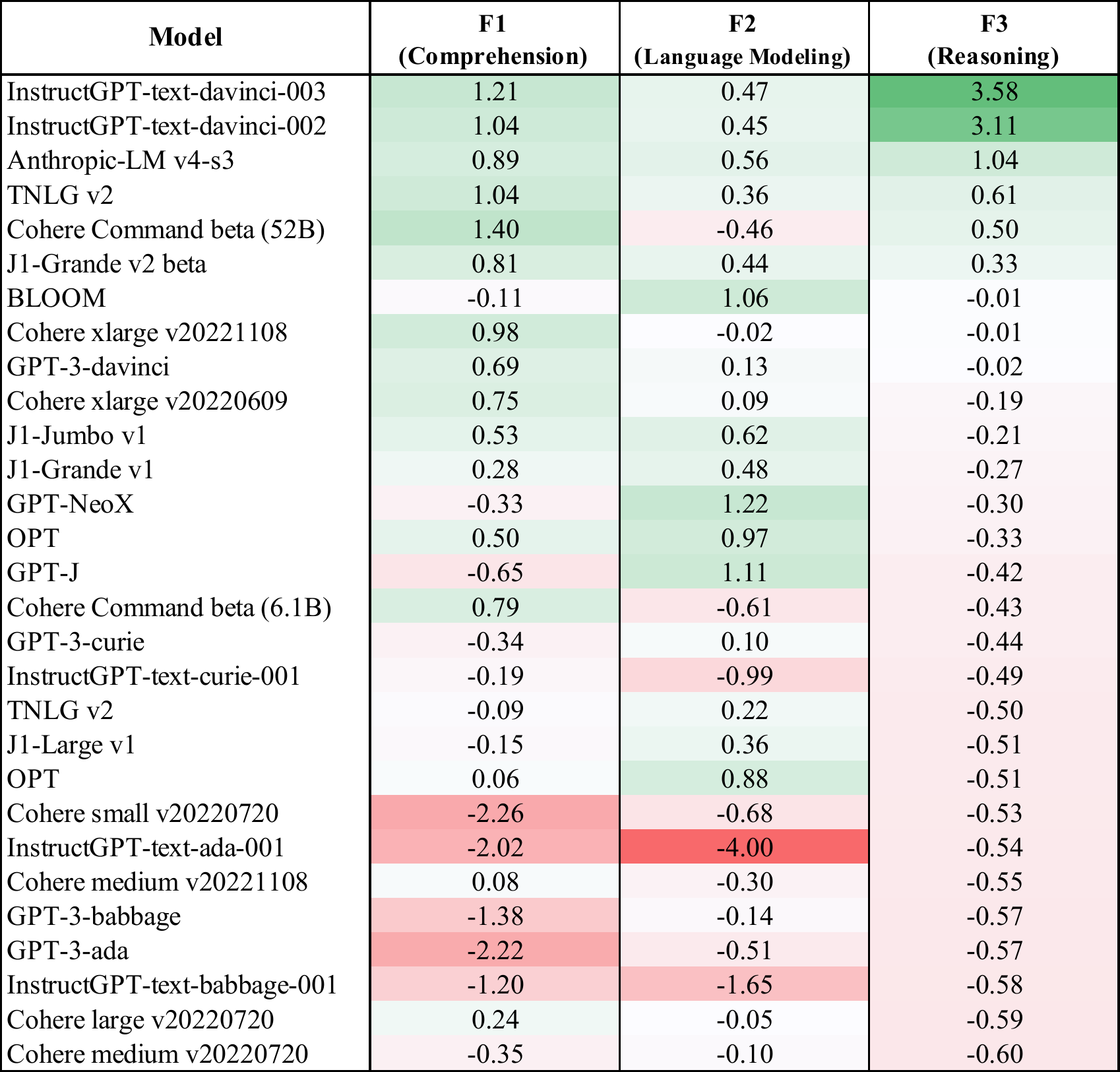}
\caption{Factor scores for each model on the three factors based on the frequentist analysis. Darker greens represent higher factor scores (greater levels of ability), while darker reds represent lower factor scores (lower levels of ability). Models are sorted by reasoning score.}
\label{fig:factorScores}
\end{figure}

Having identified three clear factors representing different cognitive abilities, we next examined the scores for each model across each factor. The results are displayed in Fig. \ref{fig:factorScores}. We find that the InstructGPT davinci models are far and away the most capable when it comes to reasoning, and also score highly on comprehension---second only to the largest Cohere Command beta model. This pattern fits with the excellent scores the InstructGPT models achieve on the HELM benchmark in general \cite{liang2022}, but the cause of this high level of ability is worth further investigation---perhaps it is because the InstructGPT models are the largest models in the dataset to have undergone instruction tuning and RLHF. In terms of language modelling, BLOOM\snJHO{BLOOM is very multilingual, and I think that some of the `language modelling' tasks are very multilingual, so that could be an explanation.}\snRyan{Actually all the language modeling tasks used only English data} and GPT-NeoX scored the highest, in contrast to their mediocre comprehension and reasoning scores. It is also striking that the InstructGPT ada and babbage models had by far the lowest language modeling score of any model, which could indicate these models are too small to effectively fine-tune without disrupting the models' latent representations.

\subsection{Relationships with model characteristics}
Now that we have identified these three latent abilities, we can start to examine how these abilities are related to different characteristics of the model---namely, model size, instruction tuning, and training length. There were no models that underwent RLHF without instruction tuning, so we did not analyse the relationships with this property. The bottom three rows of Table \ref{tab:correlations} display the correlations between the models' factor scores and various model characteristics.

\begin{figure}[h]
\centering
\includegraphics[width = 1\textwidth]{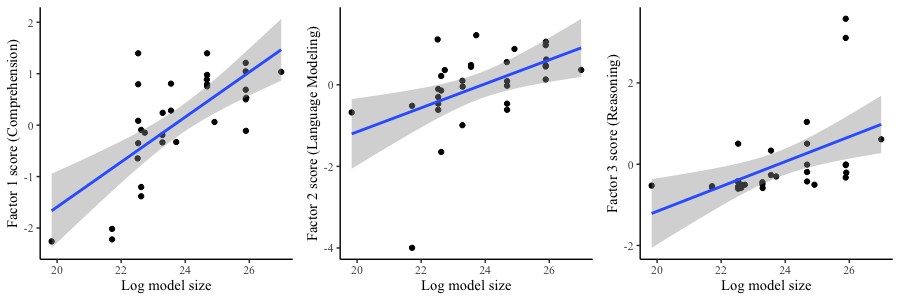}
\caption{Plots of the relationships between log model size (billions) and the extracted factor scores for each factor. Lines represent the linear relationship between the variables with  95\% confidence bands.}
\label{fig:corrs}
\end{figure}

We found that model size was positively correlated with all three factors (as a model gets bigger, it gets better at comprehension, language modeling, and reasoning). However, model size was more strongly correlated with comprehension than with either language modeling or reasoning. Plots of the relationship between model size and each factor can be found in Figure \ref{fig:corrs}. 

We also found divergence between the three abilities in terms of their relationships to instruction tuning. Instruction tuning was negatively correlated with language modeling, but positively correlated with reasoning. This pattern is logical because instruction tuning explicitly shifts a model away from the optimisation goal of modeling language and instead towards providing responses that fit the goals of the task provided by the user. The relationship with comprehension was not significantly different from zero. 

We did not find evidence of clear relationships between any of the factors and the total number of tokens the models were trained on. But this could be due to the fact that there was relatively little variation in the training length across the dataset (see Table \ref{tab:models}). To properly test these relationships, we would need data from models that vary substantially in training length, such as those used in other studies of scaling laws \cite{hoffmann2022, touvron2023}.

\section{Discussion}
Here, we investigated the structure of language model capabilities using performance data from a range of models on the HELM benchmark tasks. We found evidence that language model capabilities are multidimensional, and identified three factors that appear to correspond to abilities capturing comprehension, language modeling, and reasoning. 

To the best of our knowledge, this is the first empirical investigation into the structure of language model capabilities. Our findings suggest that a large proportion of the variance in language model performance can be explained by a small number of latent capabilities. The fact that three factors alone explained so much of the variance in performance across the wide variety of models in the dataset is striking---the included tasks test everything from question answering and summarization to sentiment analysis and various kinds of reasoning \cite{liang2022}. This pattern indicates that although LLMs vary greatly in their capabilities, they might share a common underlying cognitive structure. If this is indeed the case, we may be able to learn a lot about the capabilities of future models by examining current and past systems---a fact that might provide some hope for efforts to understand these systems in the face of breakneck progress.

The data presented here provide useful insights into the underlying abilities needed to complete different kinds of tasks. For instance, the results suggest that various tasks involving comprehension, including question answering, summarization, and sentiment analysis all appear to rely on the same underlying ability, despite having quite different formats. Similarly, our data suggest that mathematical reasoning tasks and other inductive reasoning tasks might rely on a single underlying reasoning ability.

These findings also have a number of implications for benchmark construction. If the performance of LLMs on a wide range of tasks can be largely explained by a small number of broad capabilities, it should be possible to make benchmarks more efficient. Instead of requiring dozens or even hundreds of tasks, we may be able to obtain a good sense of a model's capabilities with only a small number of tasks that provide estimates of a model's reasoning, comprehension, and language modeling abilities. This approach would provide a much more cost-efficient way of gathering useful data on a wide range of models. Such an outcome would be in-keeping with wider efforts to make more streamlined, "lite" benchmarks to combat the prohibitive costs involved in benchmarking LLMs \cite{liang2022, srivastava2022}. However, many of these streamlined benchmarks, including BIG-bench lite, take a somewhat arbitrary approach to selecting which tasks to include \cite{srivastava2022}. By contrast, in line with calls to make benchmarks more systematic \cite{burnell2023, raji2021}, our approach provides an empirically-driven way of selecting benchmark tasks based on the underlying abilities those tasks measure and the unique variance they explain. Moreover, because this approach centres around high-level cognitive abilities that are not tied to any specific task, we can use estimates of these abilities to predict performance on untested tasks, provided we know what abilities the untested tasks are likely to require. 

This study is an important first step on the road to understanding language model capabilities, but it should not be thought of as a complete accounting of every possible ability. Although the HELM tasks included in this analysis cover a wide range of the key tasks LLMs are used for, there may be other important abilities that were not tested here (e.g., abilities involved in creative tasks such as writing poetry or novels). The results of any factor analysis depend on the tasks included in the analysis, so it is important that future work examines how the factor structure changes (if at all) when other kinds of tasks are added to the mix. Indeed, another way of looking at these findings is that, despite the fact that the HELM tasks were intended to test a variety of abilities, they really only tap into the three main abilities identified here. From that perspective, our findings suggest that new task paradigms may be needed if we want to tap into abilities beyond comprehension, reasoning, and language modeling. 

These findings also further our understanding of how the properties of a model affect its capabilities. In general, we found support for scaling laws---all three of the extracted factors were positively correlated with model size. However, we also found evidence that scale is not uniformly related to the three abilities. More specifically, we found that model size was more strongly related to comprehension than it was to either reasoning or language modeling, which could partly explain some of the inconsistencies in the findings of studies investigating the relationships between model size and task performance \cite{srivastava2022, brown2020}. We also found other evidence of dissociations between the different abilities. For example, instruction tuning was negatively related to language modeling ability, but positively related to reasoning. These findings demonstrate that changes to the properties of a model can have differential effects on the different capabilities of the model---changes that improve one ability might actually impair one or more other abilities. These divergences also highlight the importance of considering the specific abilities involved in a set of tasks when attempting to draw conclusions about the effects of particular model properties on task performance.

Finally, our findings demonstrate the broader value of making evaluation data publicly available \cite{burnell2023}. Datasets containing evaluation results, such as the one released by the creators of HELM \cite{liang2022}, provide an incredibly useful tool for building a deeper understanding of AI systems, and we encourage researchers and organizations to publicly release their own benchmark data wherever possible.

There are, naturally, several limitations to our conclusions. First, given the relatively small sample of LLMs in this study, it is important to replicate these findings in a larger sample of models. To the best of our knowledge, the HELM dataset is currently the most comprehensive public dataset of benchmark data from LLMs, but given the speed at which models are being released, it should soon be possible to obtain larger samples. Additional data could also be obtained by performing ablations of different models and testing each version on the same benchmark. Second, the interpretations of the factors in any factor analysis are inherently somewhat subjective. Although there was a relatively clear divide between the reasoning and comprehension tasks, there were some tasks that did not fit neatly, suggesting that future work is needed to confirm the validity of these interpretations. Third, the tasks were annotated according to the main cognitive ability being tested by the task. But many of the tasks in HELM were designed based on the format of the task (e.g., question answering), rather than based on the cognitive demands of the task (e.g., fluid reasoning). To more cleanly separate out the capabilities of LLMs, it is important that future benchmarks include tasks specifically designed to test for particular cognitive abilities. 

Despite these limitations, our findings provide a valuable first step on the path to developing a theoretical understanding the capabilities of large language models. We hope that future work can build on this foundation, using a variety of empirical methods to identify the fundamental building blocks of language model cognition. Given the rapid advancement and widespread deployment of LLMs, we believe a full understanding of these building blocks will be crucial for our ability to predict the complex behavior of these complex prediction machines.

\newpage

\bibliographystyle{plain}
\bibliography{bib}

\begin{thebibliography}{10}

\bibitem{brown2020}
Tom~B. Brown, Benjamin Mann, Nick Ryder, Melanie Subbiah, Jared Kaplan,
  Prafulla Dhariwal, Arvind Neelakantan, Pranav Shyam, Girish Sastry, Amanda
  Askell, Sandhini Agarwal, Ariel Herbert-Voss, Gretchen Krueger, Tom Henighan,
  Rewon Child, Aditya Ramesh, Daniel~M. Ziegler, Jeffrey Wu, Clemens Winter,
  Christopher Hesse, Mark Chen, Eric Sigler, Mateusz Litwin, Scott Gray,
  Benjamin Chess, Jack Clark, Christopher Berner, Sam McCandlish, Alec Radford,
  Ilya Sutskever, and Dario Amodei.
\newblock Language {{Models}} are {{Few-Shot Learners}}.

\bibitem{burnell2022}
Ryan Burnell, John Burden, Danaja Rutar, Konstantinos Voudouris, Lucy Cheke,
  and José Hernández-Orallo.
\newblock Not a {{Number}}: {{Identifying Instance Features}} for
  {{Capability-Oriented Evaluation}}.
\newblock In {\em Proceedings of the {{Thirty-First International Joint
  Conference}} on {{Artificial Intelligence}}}, pages 2827--2835.
  {International Joint Conferences on Artificial Intelligence Organization}.

\bibitem{burnell2023}
Ryan Burnell, Wout Schellaert, John Burden, Tomer~D. Ullman, Fernando
  Martinez-Plumed, Joshua~B. Tenenbaum, Danaja Rutar, Lucy~G. Cheke, Jascha
  Sohl-Dickstein, Melanie Mitchell, Douwe Kiela, Murray Shanahan, Ellen~M.
  Voorhees, Anthony~G. Cohn, Joel~Z. Leibo, and Jose Hernandez-Orallo.
\newblock Rethink reporting of evaluation results in {{AI}}.
\newblock 380(6641):136--138.

\bibitem{carroll1997}
John~B. Carroll.
\newblock Psychometrics, intelligence, and public perception.
\newblock 24(1):25--52.

\bibitem{cattell1978}
Raymond~B. Cattell and John~L. Horn.
\newblock A check on the theory of fluid and crystallized intelligence with
  description of new subtest designs.
\newblock 15(3):139--164.

\bibitem{clark2020}
Peter Clark, Oyvind Tafjord, and Kyle Richardson.
\newblock Transformers as {{Soft Reasoners}} over {{Language}}.

\bibitem{conti2014}
Gabriella Conti, Sylvia Frühwirth-Schnatter, James~J. Heckman, and Rémi
  Piatek.
\newblock Bayesian exploratory factor analysis.
\newblock 183(1):31--57.

\bibitem{conway2015}
Andrew R.~A. Conway and Kristof Kovacs.
\newblock New and emerging models of human intelligence.
\newblock 6(5):419--426.

\bibitem{fabrigar2012}
Leandre~R. Fabrigar and Duane~T. Wegener.
\newblock {\em Exploratory Factor Analysis}.
\newblock Exploratory Factor Analysis. {Oxford University Press}.

\bibitem{flaim2020}
Mary Flaim and Aaron~P. Blaisdell.
\newblock The comparative analysis of intelligence.
\newblock 146(12):1174--1199.

\bibitem{flanagan2013}
Dawn~P. Flanagan, Vincent~C. Alfonso, and Matthew~R. Reynolds.
\newblock Broad and {{Narrow CHC Abilities Measured}} and {{Not Measured}} by
  the {{Wechsler Scales}}: {{Moving Beyond Within-Battery Factor Analysis}}.
\newblock 31(2):202--223.

\bibitem{flanagan2010}
Dawn~P. Flanagan, Catherine~A. Fiorello, and Samuel~O. Ortiz.
\newblock Enhancing practice through application of
  {{Cattell}}–{{Horn}}–{{Carroll}} theory and research: {{A}} “third
  method” approach to specific learning disability identification.
\newblock 47(7):739--760.

\bibitem{healy2007}
Susan~D Healy and Candy Rowe.
\newblock A critique of comparative studies of brain size.
\newblock 274(1609):453--464.

\bibitem{hendrycks2021}
Dan Hendrycks, Collin Burns, Steven Basart, Andy Zou, Mantas Mazeika, Dawn
  Song, and Jacob Steinhardt.
\newblock Measuring {{Massive Multitask Language Understanding}}.

\bibitem{hoffmann2022}
Jordan Hoffmann, Sebastian Borgeaud, Arthur Mensch, Elena Buchatskaya, Trevor
  Cai, Eliza Rutherford, Diego de~Las Casas, Lisa~Anne Hendricks, Johannes
  Welbl, Aidan Clark, Tom Hennigan, Eric Noland, Katie Millican, {George van
  den Driessche}, Bogdan Damoc, Aurelia Guy, Simon Osindero, Karen Simonyan,
  Erich Elsen, Jack~W. Rae, Oriol Vinyals, and Laurent Sifre.
\newblock Training {{Compute-Optimal Large Language Models}}.

\bibitem{kaufman2014}
Alan~S. Kaufman and Nadeen~L. Kaufman.
\newblock Kaufman {{Brief Intelligence Test}}, {{Second Edition}}.
\newblock In {\em Encyclopedia of {{Special Education}}}. {John Wiley \& Sons,
  Ltd}.

\bibitem{kovacs2016}
Kristof Kovacs and Andrew R.~A. Conway.
\newblock Process {{Overlap Theory}}: {{A Unified Account}} of the {{General
  Factor}} of {{Intelligence}}.
\newblock 27(3):151--177.

\bibitem{lewkowycz2022}
Aitor Lewkowycz, Anders Andreassen, David Dohan, Ethan Dyer, Henryk
  Michalewski, Vinay Ramasesh, Ambrose Slone, Cem Anil, Imanol Schlag, Theo
  Gutman-Solo, Yuhuai Wu, Behnam Neyshabur, Guy Gur-Ari, and Vedant Misra.
\newblock Solving {{Quantitative Reasoning Problems}} with {{Language Models}}.

\bibitem{liang2022}
Percy Liang, Rishi Bommasani, Tony Lee, Dimitris Tsipras, Dilara Soylu,
  Michihiro Yasunaga, Yian Zhang, Deepak Narayanan, Yuhuai Wu, Ananya Kumar,
  Benjamin Newman, Binhang Yuan, Bobby Yan, Ce~Zhang, Christian Cosgrove,
  Christopher~D. Manning, Christopher Ré, Diana Acosta-Navas, Drew~A. Hudson,
  Eric Zelikman, Esin Durmus, Faisal Ladhak, Frieda Rong, Hongyu Ren, Huaxiu
  Yao, Jue Wang, Keshav Santhanam, Laurel Orr, Lucia Zheng, Mert Yuksekgonul,
  Mirac Suzgun, Nathan Kim, Neel Guha, Niladri Chatterji, Omar Khattab, Peter
  Henderson, Qian Huang, Ryan Chi, Sang~Michael Xie, Shibani Santurkar, Surya
  Ganguli, Tatsunori Hashimoto, Thomas Icard, Tianyi Zhang, Vishrav Chaudhary,
  William Wang, Xuechen Li, Yifan Mai, Yuhui Zhang, and Yuta Koreeda.
\newblock Holistic {{Evaluation}} of {{Language Models}}.

\bibitem{logan2018}
Corina~J. Logan, Shahar Avin, Neeltje Boogert, Andrew Buskell, Fiona~R. Cross,
  Adrian Currie, Sarah Jelbert, Dieter Lukas, Rafael Mares, Ana~F. Navarrete,
  Shuichi Shigeno, and Stephen~H. Montgomery.
\newblock Beyond brain size: {{Uncovering}} the neural correlates of behavioral
  and cognitive specialization.
\newblock 13:55--89.

\bibitem{lorenzo-seva2011}
Urbano Lorenzo-Seva, Marieke~E. Timmerman, and Henk A.~L. Kiers.
\newblock The {{Hull Method}} for {{Selecting}} the {{Number}} of {{Common
  Factors}}.
\newblock 46(2):340--364.

\bibitem{maccallum1999}
Robert~C. MacCallum, Keith~F. Widaman, Shaobo Zhang, and Sehee Hong.
\newblock Sample size in factor analysis.
\newblock 4:84--99.

\bibitem{mcgrew2009}
Kevin~S. McGrew.
\newblock {{CHC}} theory and the human cognitive abilities project:
  {{Standing}} on the shoulders of the giants of psychometric intelligence
  research.
\newblock 37(1):1--10.

\bibitem{mcneish2016}
Daniel McNeish.
\newblock On {{Using Bayesian Methods}} to {{Address Small Sample Problems}}.
\newblock 23(5):750--773.

\bibitem{miller2008}
Bryan~D. Miller.
\newblock Cattell-{{Horn-Carroll}} ({{CHC}}) {{Theory-Based Assessment With
  Deaf}} and {{Hard}} of {{Hearing Children}} in the {{School Setting}}.
\newblock 152(5):459--466.

\bibitem{rajani2019}
Nazneen~Fatema Rajani, Bryan McCann, Caiming Xiong, and Richard Socher.
\newblock Explain {{Yourself}}! {{Leveraging Language Models}} for
  {{Commonsense Reasoning}}.

\bibitem{raji2021}
Inioluwa~Deborah Raji, Emily~M. Bender, Amandalynne Paullada, Emily Denton, and
  Alex Hanna.
\newblock {{AI}} and the {{Everything}} in the {{Whole Wide World Benchmark}}.

\bibitem{rosseel2012}
Yves Rosseel.
\newblock Lavaan : {{An R Package}} for {{Structural Equation Modeling}}.
\newblock 48(2).

\bibitem{schneider2012}
W.~Joel Schneider and Kevin~S. McGrew.
\newblock The {{Cattell-Horn-Carroll}} model of intelligence.
\newblock In {\em Contemporary Intellectual Assessment: {{Theories}}, Tests,
  and Issues, 3rd Ed}, pages 99--144. {The Guilford Press}.

\bibitem{srivastava2022}
Aarohi Srivastava et~al.
\newblock Beyond the {{Imitation Game}}: {{Quantifying}} and extrapolating the
  capabilities of language models.

\bibitem{sternberg2013}
Robert~J. Sternberg.
\newblock Contemporary theories of intelligence.
\newblock In {\em Handbook of Psychology: {{Educational}} Psychology, {{Vol}}.
  7, 2nd Ed}, pages 23--44. {John Wiley \& Sons, Inc.}

\bibitem{taylor2021}
J.~Eric~T. Taylor and Graham~W. Taylor.
\newblock Artificial cognition: {{How}} experimental psychology can help
  generate explainable artificial intelligence.
\newblock 28(2):454--475.

\bibitem{touvron2023}
Hugo Touvron, Thibaut Lavril, Gautier Izacard, Xavier Martinet, Marie-Anne
  Lachaux, Timothée Lacroix, Baptiste Rozière, Naman Goyal, Eric Hambro,
  Faisal Azhar, Aurelien Rodriguez, Armand Joulin, Edouard Grave, and Guillaume
  Lample.
\newblock {{LLaMA}}: {{Open}} and {{Efficient Foundation Language Models}}.

\bibitem{wei2022}
Jason Wei, Yi~Tay, Rishi Bommasani, Colin Raffel, Barret Zoph, Sebastian
  Borgeaud, Dani Yogatama, Maarten Bosma, Denny Zhou, Donald Metzler, Ed~H.
  Chi, Tatsunori Hashimoto, Oriol Vinyals, Percy Liang, Jeff Dean, and William
  Fedus.
\newblock Emergent {{Abilities}} of {{Large Language Models}}.

\bibitem{zhao2023}
Wayne~Xin Zhao, Kun Zhou, Junyi Li, Tianyi Tang, Xiaolei Wang, Yupeng Hou,
  Yingqian Min, Beichen Zhang, Junjie Zhang, Zican Dong, Yifan Du, Chen Yang,
  Yushuo Chen, Zhipeng Chen, Jinhao Jiang, Ruiyang Ren, Yifan Li, Xinyu Tang,
  Zikang Liu, Peiyu Liu, Jian-Yun Nie, and Ji-Rong Wen.
\newblock A {{Survey}} of {{Large Language Models}}.

\end{thebibliography}

\appendix
\newpage
\section*{Appendix}\label{appendix}

\section{Correlations between tasks}\label{appendixA}

\begin{sidewaysfigure}[!htb]
\includegraphics{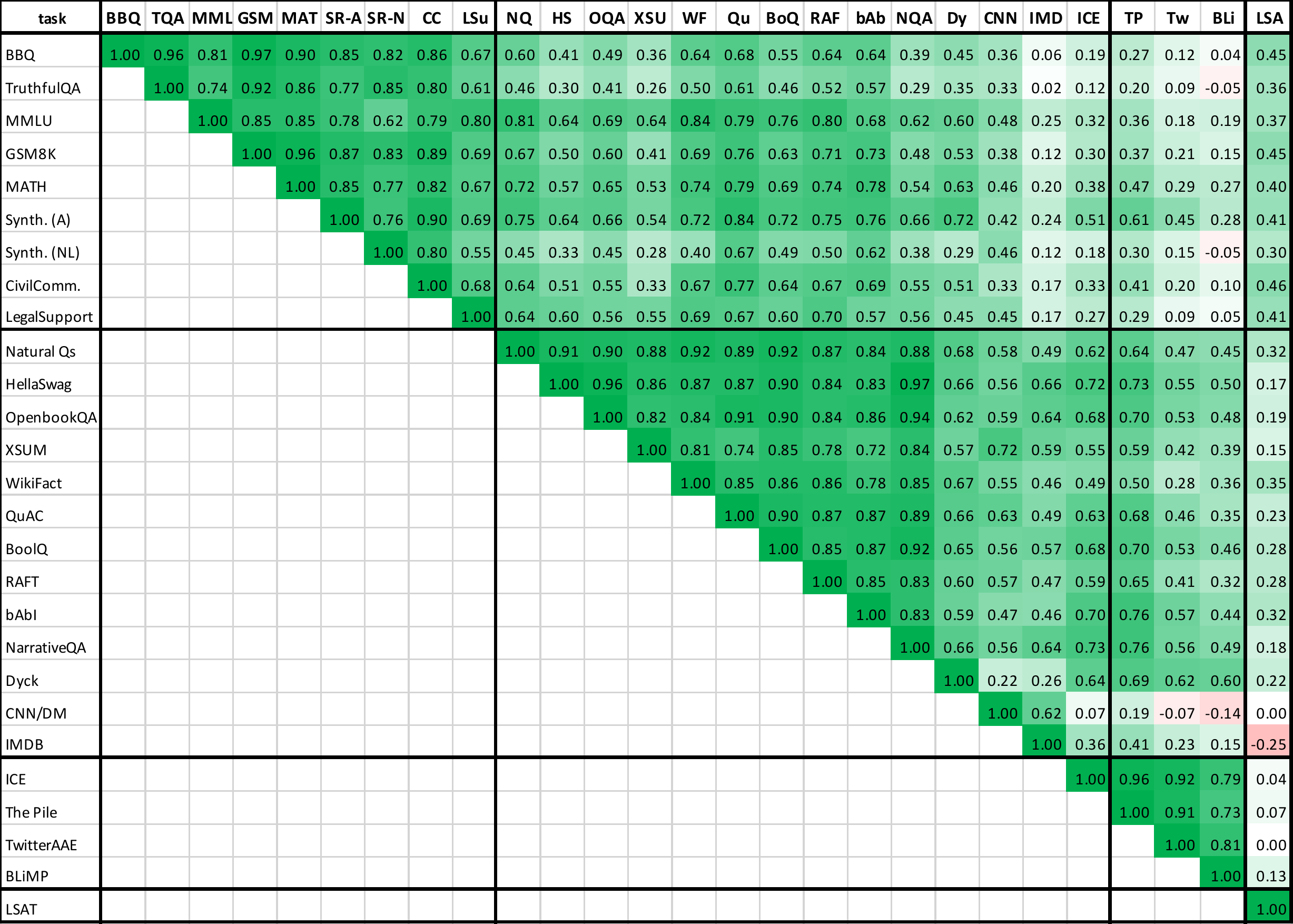}
    \caption{Pearson correlations between tasks (r). Darker greens represent stronger positive correlations, darker reds represent stronger negative correlations. Lines separate tasks according to the assigned factors in the Bayesian factor analysis.}
    \label{fig:corrsTab}
\end{sidewaysfigure}

As an initial effort to examine how the HELM tasks are related to one another, we calculated Pearson correlations between each task. These correlations are displayed in Figure \ref{fig:corrsTab}. Note that column names are abbreviated to save space---the order corresponds to the row order.

\clearpage

\section{Determining Factor Structure}\label{appendixB}

\begin{figure}[!htb]
\centering
\includegraphics[width = 1\textwidth]{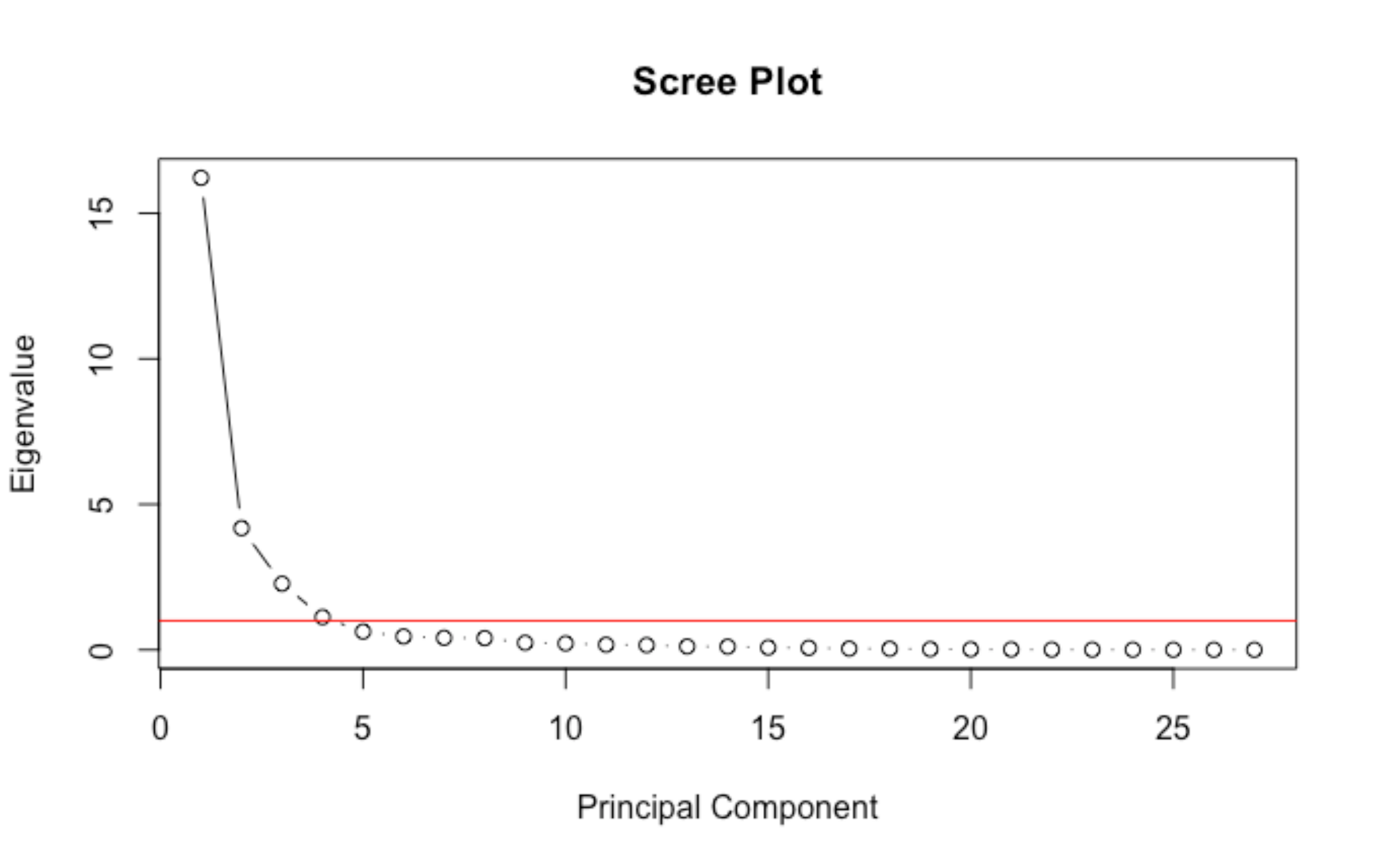}
\caption{Scree plot of eigenvalues associated with each factor. The red line represents the standard cutoff (eigenvalue = 1).}
\label{fig:screePlot}
\end{figure}

We conducted a number of steps to determine the factor structure of the data. First, we analyzed the scree plot and examined the eigenvalues of each extracted factor (see Figure \ref{fig:screePlot}). The patterns suggest a 3 or 4-factor solution is most appropriate

\begin{figure}[!h]
\centering
\includegraphics[width = 1\textwidth]{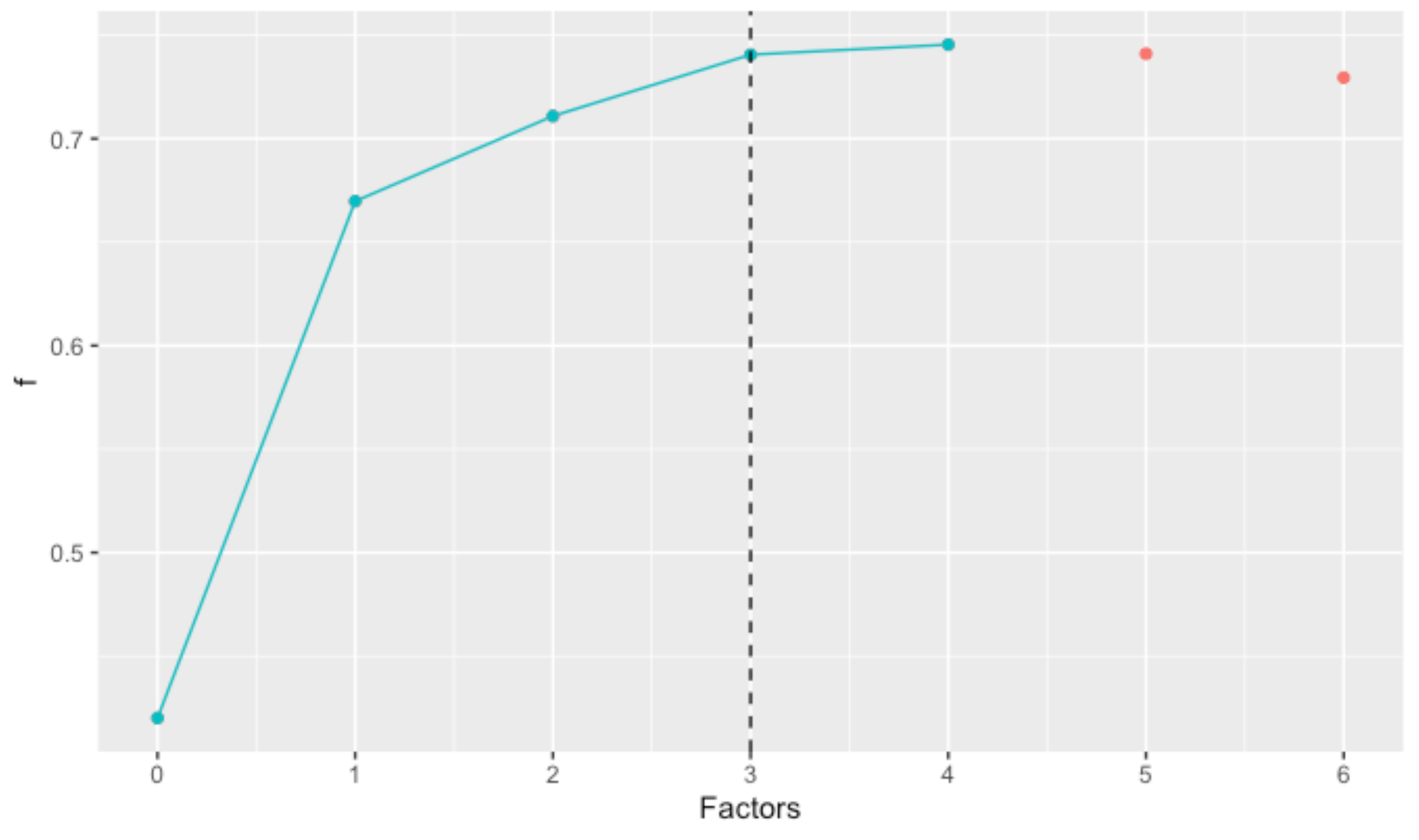}
\caption{Hull method plot of goodness of fit (f; calculated as 1 - RMSEA), against model degrees of freedom.}
\label{fig:hullPlot}
\end{figure}

Next, we employed a more systematic method of determining the factor structure---the Hull method \cite{lorenzo-seva2011}, which aims to balance the goodness of fit against the degrees of freedom of the model. As shown in Figure \ref{fig:hullPlot}, this method suggests a 3-factor solution is most appropriate (determining that the potential fourth factor would not improve goodness of fit sufficiently to warrant a more complex model).

\begin{figure}[!h]
\centering
\includegraphics[width = 1\textwidth]{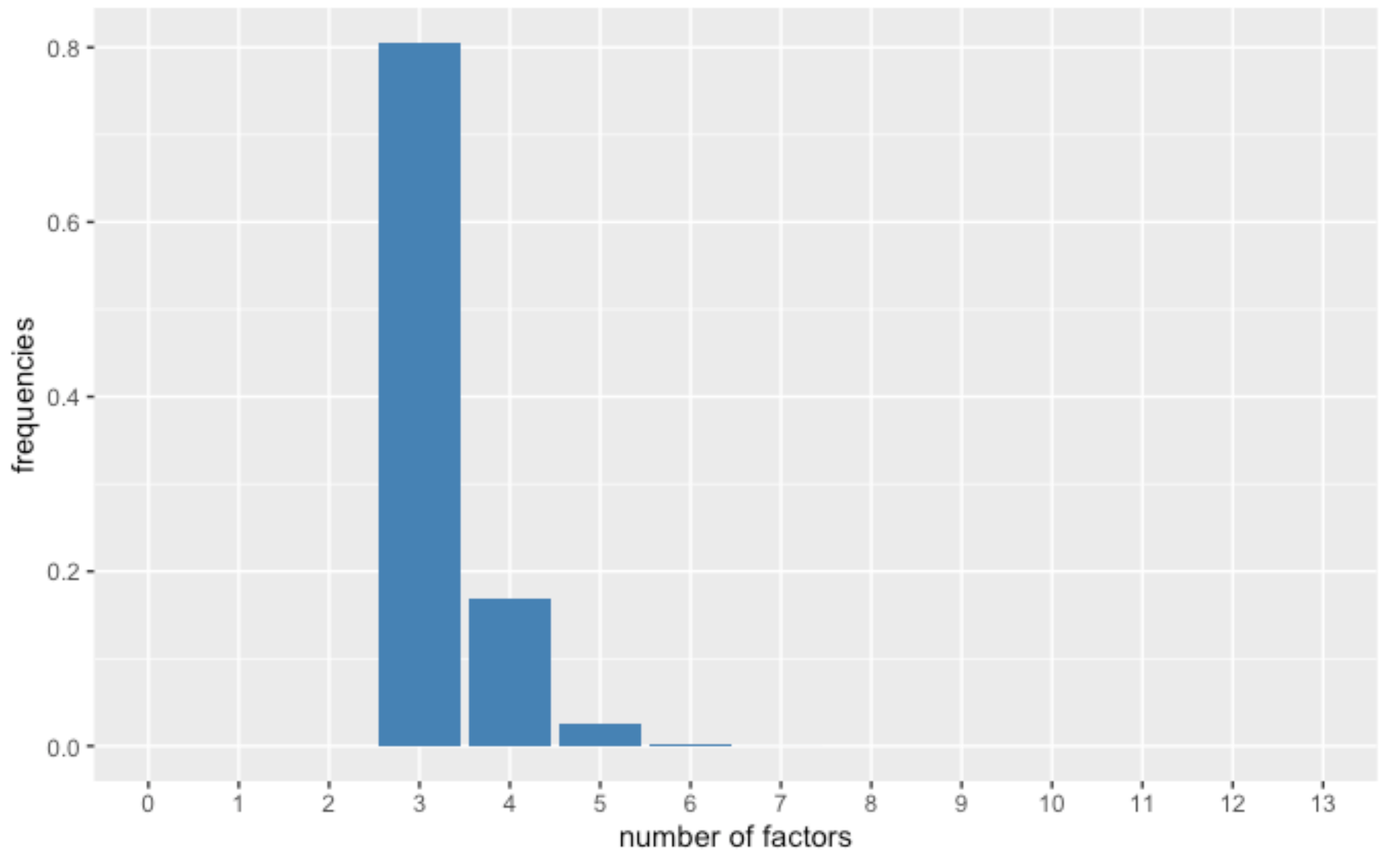}
\caption{Posterior distribution of the number of factors based on the Bayesian Factor Analysis.}
\label{fig:bayesNumFactors}
\end{figure}

These findings align with the Bayesian factor analysis, which determines the most likely number of factors as part of the modelling process. As shown in the posterior distribution displayed in Figure \ref{fig:bayesNumFactors}, the analysis determined that it is most likely that the data are explained by three underlying factors. Based on this converging evidence, we conducted the final frequentist exploratory factor analysis with 3 factors.

\end{document}